\documentclass[journal]{IEEEtran}
\usepackage{amsmath,amsfonts}
\usepackage{algorithmic}
\usepackage{algorithm}
\usepackage{array}
\usepackage[caption=false,font=normalsize,labelfont=sf,textfont=sf]{subfig}
\usepackage{textcomp}
\usepackage{stfloats}
\usepackage{url}
\usepackage{verbatim}
\usepackage{graphicx}
\usepackage{cite}
\usepackage{hyperref} 

\usepackage{array}
\usepackage{multirow}
\usepackage{longtable}
\usepackage{rotating}
\usepackage{float}
\usepackage{booktabs}
\usepackage{tabularx}
\usepackage{ulem}
\usepackage{bbding}
\usepackage{makecell}
\usepackage[switch]{lineno}

\usepackage{tikz,xcolor,hyperref}
\definecolor{lime}{HTML}{A6CE39}
\DeclareRobustCommand{\orcidicon}{%
    \begin{tikzpicture}
    \draw[lime, fill=lime] (0,0) 
    circle [radius=0.16] 
    node[white] {{\fontfamily{qag}\selectfont \tiny ID}};    \draw[white, fill=white] (-0.0625,0.095) 
    circle [radius=0.007];    \end{tikzpicture}
    \hspace{-2mm}}
\foreach \x in {A, ..., Z}{%
    \expandafter\xdef\csname orcid\x\endcsname{\noexpand\href{https://orcid.org/\csname orcidauthor\x\endcsname}{\noexpand\orcidicon}}
    }

\hypersetup{
	colorlinks=true,
	linkcolor=red,
	filecolor=blue,      
	urlcolor=red,
	citecolor=green,
}

\begin{document}


\title{Double-chain Constraints for 3D Human Pose Estimation in Images and Videos}


\author{\IEEEmembership{Hongbo~Kang, Yong~Wang, Mengyuan Liu, Doudou~Wu, Peng~Liu, Wenming~Yang, Senior Member, IEEE}
 
\thanks{This work was partly supported by the National Natural Science Foundation of China(No.62171251) and the Special Foundations for the Development of Strategic Emerging Industries of Shenzhen(Nos.JCYJ20200109143035495, CJGJZD20210408092804011 \& JSGG20211108092812020).}
\thanks{H. Kang and Y. Wang are equally-contributed authors.}
\thanks{Yong Wang is the corresponding author. Email: ywang@cqut.edu.cn}
\thanks{H. Kang, Y. Wang, D. Wu, P. Liu are with the School of Artificial Intelligence,
Chongqing University of Technology, China.}
\thanks{M. Liu is with the Shenzhen Graduate School, Peking University, China.}
\thanks{W. Yang is with the Shenzhen International Graduate School/Department of Electronic Engineering, Tsinghua University, China.}}





\maketitle

\begin{abstract}


Reconstructing 3D poses from 2D poses lacking depth information is particularly challenging due to the complexity and diversity of human motion. The key is to effectively model the spatial constraints between joints to leverage their inherent dependencies.  Thus, we propose a novel model, called Double-chain Graph Convolutional Transformer (DC-GCT), to constrain the pose through a double-chain design consisting of local-to-global and global-to-local chains to obtain a complex representation more suitable for the current human pose. Specifically, we combine the advantages of GCN and Transformer and design a Local Constraint Module (LCM) based on GCN and a Global Constraint Module (GCM) based on self-attention mechanism as well as a Feature Interaction Module (FIM). The proposed method fully captures the multi-level dependencies between human body joints to optimize the modeling capability of the model. Moreover, we propose a method to use temporal information into the single-frame model by guiding the video sequence embedding through the joint embedding of the target frame, with negligible increase in computational cost. Experimental results demonstrate that DC-GCT achieves state-of-the-art performance on two challenging datasets (Human3.6M and MPI-INF-3DHP). Notably, our model achieves state-of-the-art performance on all action categories in the Human3.6M dataset using detected 2D poses from CPN, and our code is available at: \href{https://github.com/KHB1698/DC-GCT}{https://github.com/KHB1698/DC-GCT}.

\end{abstract}


\begin{figure}[t]
\centering
\includegraphics[width=\linewidth]{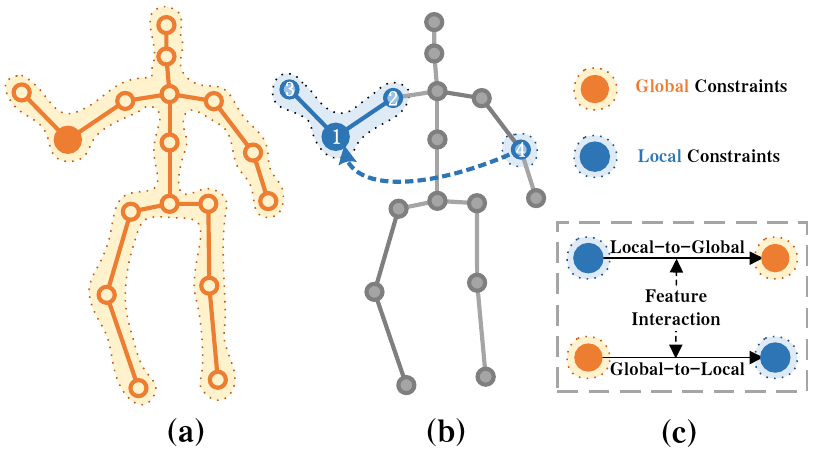}
\caption{Overview of local and global constraints in our double-chain structure. (a) The global constraints on the right elbow that originate from all joints; (b) The local constraints on the right elbow that originate from four adjacent joints; (c) Our double-chain constraints imposes both local-to-global and global-to-local constraints on each node through separate pipelines, with feature interaction occurring at the transition stage.}
\label{fig:l_g}
\end{figure}

\section{Introduction}
\IEEEPARstart{3D}{} human pose estimation (3DHPE) plays a critical role in various applications, such as action recognition \cite{liu2017enhanced,liu2018recognizing,wang2018depth}, human computer interaction \cite{preece1994human}, augmented reality (AR) and virtual reality (VR) \cite{mehta2017vnect}. The primary objective of 3DHPE is to recover the 3D positions of human joints from images or videos. This can be achieved through two distinct methods: direct estimation\cite{zhou2016deep,lee2018propagating,chen2020towards} and 2D-to-3D lifting \cite{zhao2022graformer,pavllo20193d,cai2019exploiting,hu2021conditional}. The method of 2D-to-3D lifting requires the estimation of 2D pose from a single image or video, followed by the conversion to 3D joint coordinates of the pose. The superior performance of recent advanced 2D pose detectors \cite{chen2018cascaded,sun2019deep} has made the 2D-to-3D lifting methods more effective than direct estimation methods. However, estimating 3D pose from a 2D representation is an ill-posed problem since there may exist multiple valid 3D interpretations for a single 2D skeleton.

One feasible solution is to utilize temporal information from multiple frames to improve the estimation accuracy. However, the multi-frame approach \cite{pavllo20193d,zheng20213d,zhao2023poseformerv2} requires high computational costs. In contrast, the single-frame approach is more suitable for providing efficient solutions in scenarios with limited computational resources and tight time requirements. Additionally, the single-frame approach can also handle the problem of human pose estimation in static images, providing an important foundation for image-based human pose estimation research. Therefore, developing a powerful single-frame 3D pose lifter can not only improve the performance for single image but also provide more efficient solutions for video data. However, single-frame tasks present greater challenges than multi-frame tasks because of the limited information that can be obtained from a single 2D pose. This fundamental challenge motivates our work, in which we propose a powerful solution. Furthermore, we explore a feasible method to embed multi-frame information into a single-frame model. This approach obtains additional temporal information as input while ensuring a relatively small increase in computational cost.

In recent years, some works \cite{cai2019exploiting,xu2021graph,liu2020comprehensive,zhao2019semantic,zou2021modulated} have explored the use of graph convolutional networks (GCNs) for modeling the correlations and local dependency relationships between body joints in human skeletons, which are naturally represented as graphs. While GCNs have demonstrated good performance on skeleton data, they have some limitations. When building deeper GCN-based models, the problem of over-smoothing often arises. This means that as the number of layers increases, the representations of adjacent nodes become increasingly similar, resulting in information loss and confusion between nodes. To avoid this problem, most GCN-based models use fewer layers. However, having fewer layers may limit the model's ability to capture global dependencies between nodes. Recently, transformer-based models \cite{vaswani2017attention,carion2020end,bao2021beit} in computer vision have shown high performance on various tasks. In Transformer-based methods\cite{zheng20213d,lutz2022jointformer,zhao2022graformer}, the self-attention mechanism is employed to connect the representation of each node with those of other nodes, reducing confusion among nodes and capturing global dependencies between them. However, the self-attention mechanism primarily focuses on joint similarity while weakening or disregarding structural information in 2D joint coordinates. Thus, it is essential to devise a model that can efficiently integrate the benefits of GCN and transformer to capture local and global dependency relationships between joints in human skeletons.

\begin{figure*}[!t]
  \centering
  \includegraphics[width=\textwidth]{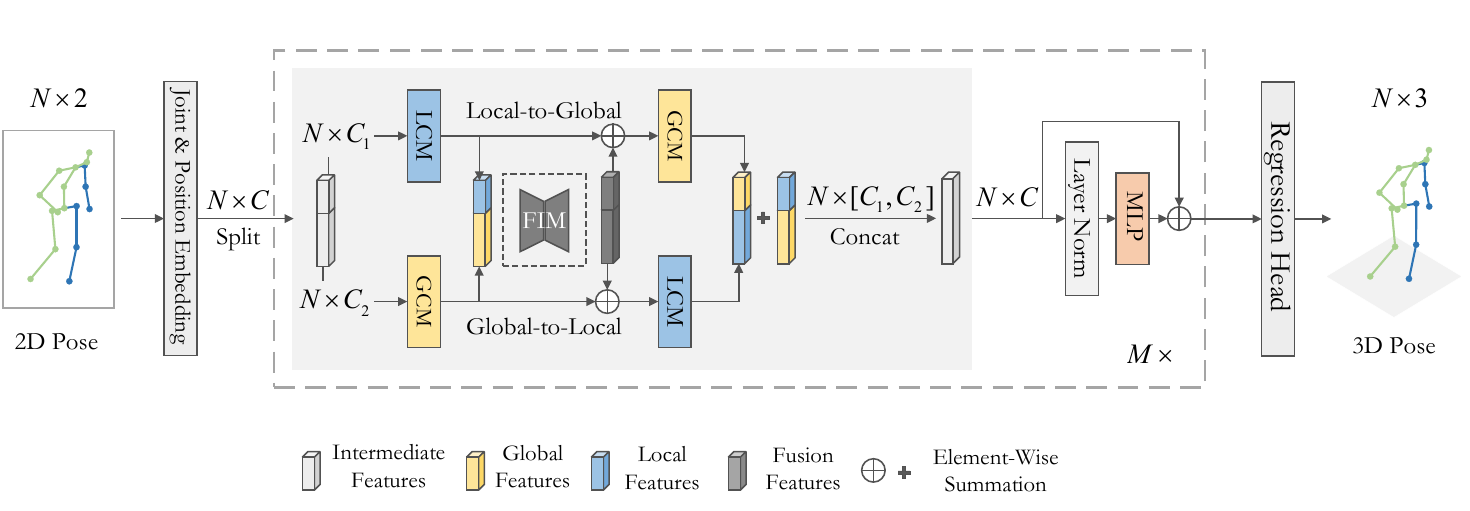}
  \caption{Overview of the proposed Double-chain Graph Convolution Transformer (DC-GCT). The DC-GCT consists of $M$ stacked double-chain structures, where each structure splits the channel dimension $C$ into $C_1$ and $C_2$, and applies local-to-global and global-to-local constraints to capture the dependencies of human pose, respectively.}
  \label{fig:structure1}
\end{figure*}

Due to the complex and dynamic nature of human motion, an accurate 3D representation of the human body requires a method that captures the multi-level interdependencies of the human body from 2D poses. Therefore, the proposed model should accomplish this goal by applying both local and global constraints to different human joints. In our work,  the global constraints (Figure\ref{fig:l_g}(a)) originate from feature constraints imposed by all nodes on the target node, while the local constraints (Figure\ref{fig:l_g}(b)) originate from feature constraints imposed by physically neighboring and indirectly adjacent (symmetric) nodes on the target node, and the double-chain constraints (Figure \ref{fig:l_g}(c)) imposes local-to-global and global-to-local constraints on each node. To this end, we propose the Double-chain Graph Convolutional Transformer (DC-GCT) shown in Figure\ref{fig:structure1} to capture the multi-level dependencies of the human body by implementing different constraints. Specifically, we designed a Local Constraint Module (LCM) based on GCN, which employs the GCN module to model the constraint information of different types of adjacent nodes on the target node. We also designed a Global Constraint Module (GCM) based on the self-attention mechanism of Transformer. This module learns the similarity between nodes through query, key and value matrices to model the constraint information of all nodes on the target node. We also designed a Feature Interaction Module (FIM) to automatically fuse and map local and global features. By combining the LCM, GCM and FIM, our model implements double-chain constraints to learn complex representations that are better suited for the current human pose. Moreover, we propose a method to use temporal information into the single-frame model by guiding the video sequence embedding through the joint embedding of the target frame, with negligible increase in computational cost.

We conducted comprehensive evaluations and ablation studies on two benchmark datasets (Human3.6M \cite{ionescu2013human3} and MPI-INF-3DHP \cite{mehta2017monocular}) to demonstrate the effectiveness and generalization ability of the proposed method. Experimental results show that we achieved state-of-the-art performance. We obtained first-place results in all action categories of the Human3.6M dataset using detected 2D poses from CPN. Our contributions can be summarized as follows:

\begin{itemize}
  \item To address the limitations of previous work on modeling human pose under spatial constraints, we propose modeling the double-chain (local-to-global and global-to-local) constraints between joints to capture the hierarchical dependencies in human poses.

  \item This paper introduces a novel double-chain model called DC-GCT. The proposed model combines three modules: the LCM for local constraints, the GCM for global constraints, and the FIM for feature interaction. By combining these three modules, we achieved the proposed double-chain constraints.
  \item We propose a method to embed extra temporal information into our single-frame model with negligible increase in computational cost.

  \item Our DC-GCT achieves state-of-the-art performance on two challenging datasets for 3DHPE, and has excellent generalization ability.

\end{itemize}

\section{Related Work}

\subsection{3D Human Pose Estimation}

3D human pose estimation can be categorized into two approaches: direct estimation and 2D to 3D lifting. Direct estimation methods infer 3D human pose from 2D images without intermediate steps in estimating 2D human pose representation, \textit{e.g.} \cite{pavlakos2017coarse,sun2018integral,zhao2019semantic,liu2019feature}. In contrast, 2D-to-3D lifting approaches estimate 2D human pose first before inferring 3D human pose. With the rapid development of 2D human pose estimation algorithms \cite{fang2017rmpe,cao2017realtime,chen2018cascaded,sun2019deep}, the second approach also shows better performance.  Martinez \textit{et al.} \cite{martinez2017simple} proposed a fully connected residual network for 2D-to-3D human pose estimation, which showed promising outcomes in relevant benchmark tests. Chen \textit{et al.} \cite{chen20173d} proposed using pose matching to address the issue of 2D-to-3D lifting. While some studies improve performance by utilizing temporal information in videos, such as Pavllo \textit{et al.}\cite{pavllo20193d} proposed a temporal convolutional network to incorporate time information in 2D pose sequences and Zheng \textit{et al.} \cite{zheng20213d} proposed using transformers to capture long-term spatiotemporal dependencies. But multi-frame methods are not suitable for real-time applications since they require longer sequence predictions. Therefore, our research primarily focuses on single-frame lifting methods for real-time application scenarios, while also discussing the transferability of proposed methods with temporal information.

\subsection{Graph Convolutional Network}

GCN \cite{kipf2016semi} utilizes convolutional operations to propagate information through the graph structure in order to process graph data, and has shown promising results in various applications including social network analysis \cite{wu2018socialgcn}, recommendation systems, and action recognition \cite{yan2018spatial,zhang2019graph,chen2021channel}. As human joint physical topology is a graph structure, some works have utilized graph convolutional networks to solve the 2D to 3D pose estimation problem and achieved state-of-the-art results. For example, LCN \cite{ci2019optimizing} introduced local connection networks to enhance the representation ability of GCN. SemGCN \cite{zhao2019semantic} proposed an improved graph convolution operation to learn the semantic relationships between human joints. MGCN \cite{zou2021modulated} improved SemGCN by introducing weight modulation and affinity modulation. GraphSH\cite{xu2021graph} proposes a  graph stacked hourglass  architectures to learn local and global feature representations of the human skeleton. RS-Net\cite{hassan2023regular} proposes the use of matrix splitting in conjunction with weight and adjacency modulation. The above works are to improve the internal or overall architecture of GCN, but the receptive field of the single-layer graph convolutional neural network is limited, and it is more suitable to capture the local dependency relationships and structural information of human poses.


\subsection{Transformer and Self-Attention Mechanism}


Transformer \cite{vaswani2017attention}, which captures long-term dependencies in sequence data using self-attention mechanisms, has received widespread attention in various fields. Therefore, many methods based on transformer have been proposed to solve visual tasks, such as image classification \cite{dosovitskiy2020image}, object detection \cite{carion2020end}, semantic segmentation \cite{zheng2021rethinking}, etc. In addition, many Transformer-based 3D human pose estimation methods \cite{zheng20213d,zhang2022mixste,shan2022p,zhao2023poseformerv2} have emerged and achieved state-of-the-art performance.

However, transformer focuses on the similarity of all joints and ignores the structural information between human joints. Therefore, some works have begun exploring the possibility of combining GCN and Transformer's self-attention mechanism to obtain better performance. For instance, Graformer \cite{zhao2022graformer} further enhanced the interaction between 2D joints to utilize their relationships. HTNet \cite{cai2023htnet} sequentially learned the structural priors of human topology from multiple semantic levels such as joints, parts, and body. However, the information flow in these methods is unidirectional, either from global to local or from local to global, without considering the differences between the two designs, which may lead to loss of information. Hence, our approach incorporates a double-chain architecture to capture dual-stream data concurrently, thereby enhancing performance.

\section{Approach}

The DC-GCT framework is presented in Figure \ref{fig:structure1}. In line with previous methods for lifting 2D-to-3D poses \cite{martinez2017simple,zhao2019semantic}, our approach utilizes an off-the-shelf 2D pose model to generate 2D joint coordinates $X \in \mathbb{R}^{N \times 2}$ from images. We then use these predictions to estimate the corresponding 3D joint coordinates $\widehat{Y} \in \mathbb{R}^{N \times 3}$, where $N$ represents the number of joints. Our model utilizes the LCM, GCM, and FIM modules to respectively impose local constraint, global constraint, and feature interaction, and combines them to impose double-chain constraint, enabling us to capture multi-level dependencies of the human body. The following section provides a detailed description.

\begin{figure}[t]
  \centering
  \includegraphics[width=.7\linewidth]{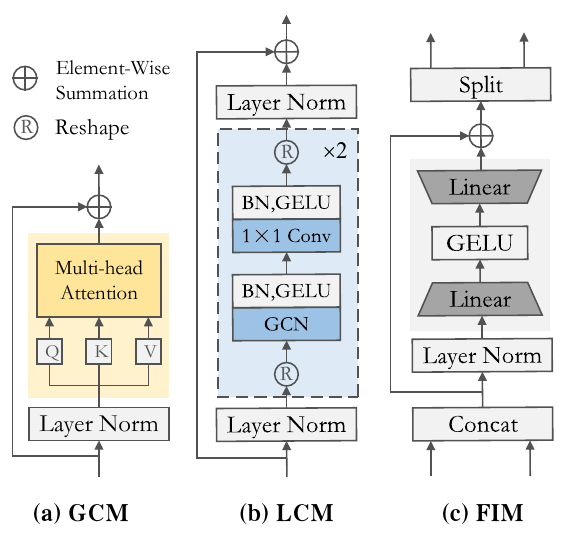}
  \caption{Overview of the components of DC-GCT. (a) Global Constraint Module. (b) Local Constraint Module.  (c) Feature Interaction Module.}
  \label{fig:structure2}
\end{figure}

\subsection{Global Constraint}

{\bf{Multi-head Self-Attention (MSA).}} The core mechanism of Transformer is MSA \cite{vaswani2017attention}. It implicitly associates input patches in the form of attention scores. The input is a query matrix $Q$, key matrix $K$ and value matrix $V$. $Q,K,V \in \mathbb{R}^{N \times C_\eta }$, where $N$ is the number of input patches, and $C_\eta $ is the dimension of each patch. The output of MSA with $h$ heads is computed as follows:
\begin{gather}
\label{eq:msa}
MSA(Q,K,V) = Concat(head_1,...,head_h)W^O
\end{gather}
where the linear projection weight is $W^O \in \mathbb{R}^{C_\eta  \times C_\eta }$ and each $head_i$ is defined as follows:
\begin{gather}
\label{eq:head}
head_i = softmax(\frac{Q_iK_i^T}{\sqrt{C_\eta'}})V_i,i\in \{1,\dots,h\}
\end{gather}
where $Q_i,K_i,V_i \in \mathbb{R}^{N \times C_\eta'}$ are the query, key, and value matrices for the $i$-th head, and $C_\eta'=C_\eta/h$ is the dimension of each head.

{\bf{Global Constraint Module (GCM).}} The feature representation of the current joint needs to take into account the constraints of other joint features and its own features, i.e., the global constraint. As the model needs to perform this operation on each joint, our GCM, as shown in Figure \ref{fig:structure2} (a), is based on MSA that aims to capture global dependency information of the human body by mutually constraining the features of all joints. The input of this module is the intermediate feature of $N$ joints, with each joint's intermediate feature treated as an independent patch, and the GCM outputs the feature representation of each joint relative to other joints. Therefore, the GCM module can effectively capture global dependency information and improve the modeling accuracy of human posture. We denote this module as $GCM(\cdot )$.

\subsection{Local Constraint}
{\bf{Graph Convolutional Network (GCN).}} The GCN receives the node features of a graph as input, characterizes node connectivity via an adjacency matrix, and employs the adjacency matrix as a convolution kernel to filter the input graph and generate novel feature representations for every node. The one-layer GCN can be expressed as follows:
\begin{gather}
\label{eq:gcn}
H' = \widetilde{D} ^{-\frac{1}{2}}\widetilde{A} \widetilde{D} ^{-\frac{1}{2}}HW
\end{gather}
where $H \in \mathbb{R}^{N \times C_\mu }$ and $H' \in \mathbb{R}^{N \times C_\nu }$ are represented the input and output feature matrix of GCN, $N$ is the number of nodes, while $C_\mu $ and $C_\nu $ represent the input and output channel dimensions. $W \in \mathbb{R}^{C_\mu  \times C_\nu }$ is the matrix of filter parameters, $\widetilde{A} = A + I_N$ is the adjacency matrix $A\in \{0,1\}^{N\times N}$ added with a diagonal matrix $I_N$ formed by self-connections. $\widetilde{D}_{ii} = \sum_j \widetilde{A}_{ij}$ is the degree matrix.

The GCN block design principle utilized in the LCM is similar to that of \cite{cai2019exploiting}. This method performs feature transformation based on adjacent nodes with different semantics and aggregates the transformed features. We only consider the spatial semantic information of 1-hop adjacent nodes and divide them into four categories: 1) the target node itself; 2) an adjacent node connected physically closer to the root node than the target node; 3) an adjacent node connected physically farther from the root node than the target node; and 4) an indirectly "symmetrically related" adjacent node. Our GCN block formula is given as follows:
\begin{gather}
\label{eq:gcn1}
H' = \sum_{k}D_k^{-\frac{1}{2}}A_kD_k^{-\frac{1}{2}}HW_k
\end{gather}
where $k$ is the index of four categories, and $W_k$ is the filter matrix for the k-th type 1-hop adjacent nodes. $A_k$ and $D_k$ are the adjacency matrix and degree matrix of the k-th type, respectively.

{\bf{Local Constraint Module (LCM).}} Due to the graphical nature of the human skeletal structure, our LCM, as shown in Figure \ref{fig:structure2} (b), is based on GCN. The aim is to constrain the feature representation of the target joint through the features of its different adjacent joints, and capture the local dependencies between joints through such local constraints. Our LCM consists of two stacks of the GCN block and the Conv block order. Layer normalization, batch normalization, and GELU operations are utilized in the middle layers to improve the effectiveness of feature representation. LCM utilizes GCN blocks to implement local constraints and vary dimensions. It then employs 1×1 Conv blocks to perform pointwise convolution and screen useful node features of high or low dimensions. We denote this module as $LCM(\cdot )$.

\subsection{Feature Interaction}

{\bf{Feature Interaction Module (FIM).}} For better the fusion and mapping of local and global features obtained from the constraint modules, we propose a FIM module, as shown in Figure \ref{fig:structure2}(c). FIM is composed of two linear layers. First, the local and global features are combined through channel concatenation and used as input. Next, linear layers are applied to reduce and recover the channel dimension to enable feature fusion and interaction. This module automatically captures interaction information between local and global features for feature fusion and determines the most valuable features to be mapped for use in the double-chain. We denote this module as $FIM_{1,2}(\cdot,\cdot)$.

\subsection{Double-chain Constraint}

{\bf{Network Architecture.}} The use of either local or global constraints alone has certain limitations. To more effectively capture the hierarchical dependencies present in human pose, we propose enhancing the model's understanding and representation by using double-chain constraints. Our approach imposes both local-to-global and global-to-local constraints on each node to capture the dependency relationships of multi-level in human pose. As shown in Figure \ref{fig:structure1}, the proposed DC-GCT is constructed by combining LCM, GCM, and FIM to achieve double-chain constraint. Below we describe the network architecture of DC-GCT in detail.

To begin, we linearly map the 2D joint coordinates $X \in \mathbb{R}^{N \times 2}$ to a high-dimensional space through joint embedding to obtain $X_0 \in \mathbb{R}^{N \times C}$, where $N$ represents the number of joints, and $C$ denotes the number of channels. Additionally, we incorporate a position matrix $E_{pos} \in \mathbb{R}^{N \times C}$, which is learnable. The position embedding can be expressed as follows:
\begin{gather}
\label{eq:embed}
X_{1} = X_{0} + E_{pos}
\end{gather}
where $X_1 \in \mathbb{R}^{N \times C}$ represents the intermediate feature representation after position embedding. This representation serves as the input to the first layer of the double-chain structure. 

The double-chain structure first splits the input features $X_m \in \mathbb{R}^{N \times C}$ of the $m$-th layer into $X_{l2g}\in \mathbb{R}^{N \times C_1}$ and $X_{g2l}\in \mathbb{R}^{N \times C_2}$ with different dimensions as follows:
\begin{gather}
\label{eq:split}
X_{l2g},X_{g2l} = Split(X_m)
\end{gather}
The resulting $X_{l2g}$ and $X_{g2l}$ are then used as inputs to the local-to-global and global-to-local chains, respectively, as follows:
\begin{gather}
\label{eq:doublechain1}
X_{l2g}^\ell=LCM(X_{l2g})  \\
X_{g2l}^\delta=GCM(X_{g2l})
\end{gather}
where $X_{l2g}^\ell \in \mathbb{R}^{N \times C_1}$  and $X_{g2l}^\delta \in \mathbb{R}^{N \times C_2}$ are the local features and global features of the double-chain, respectively. Then, the local features $X_{l2g}$ and the global features $X_{g2l}$ are automatically fused and mapped to the double-chain via FIM, as follows:
\begin{gather}
\label{eq:doublechain2}
X_{l2g}^o=X_{l2g}^\ell + FIM_1(X_{l2g}^\ell,X_{g2l}^\delta)\\
X_{g2l}^o=X_{g2l}^\delta + FIM_2(X_{l2g}^\ell,X_{g2l}^\delta)
\end{gather}
where $FIM_1$ and $FIM_2$ are the different outputs of FIM. $X_{l2g}^o \in \mathbb{R}^{N \times C_1}$  and $X_{g2l}^o \in \mathbb{R}^{N \times C_2}$ are the fused intermediate features, and then used as inputs to the local-to-global chain's GCM and global-to-local chain's LCM, respectively, as follows:
\begin{gather}
\label{eq:doublechain3}
X_{l2g}^\delta = GCM(X_{l2g}^o)\\
X_{g2l}^\ell = LCM(X_{g2l}^o)
\end{gather}
where $X_{l2g}^\delta \in \mathbb{R}^{N \times C_1}$  and $X_{g2l}^\ell \in \mathbb{R}^{N \times C_2}$ are the global features and local features of the double-chain, respectively. Finally, the local and global features of the two stages in the double-chain structure are concatenated to obtain $X_{m}'\in \mathbb{R}^{N \times C}$ as follows:
\begin{gather}
\label{eq:concat}
X_m' = Concat(X_{l2g}^\delta,X_{g2l}^\ell) + Concat(X_{l2g}^\ell,X_{g2l}^\delta)
\end{gather}

Furthermore, the double-chain structure is combined with other layers such as layer normalization (LN) and multi-layer perceptron (MLP), and the features of the double-chain are mapped as inputs to the next layer, which is defined as follows:
\begin{gather}
\label{eq:mlp}
X_{m+1} = MLP(LN(X_m')) + X_m'
\end{gather}
where $X_{m+1} \in \mathbb{R}^{N \times C}$ is the output of the $m$-th layer and the input of the $(m+1)$-th layer. $MLP$ is a multi-layer perceptron with two linear layers and a GELU activation function. $LN$ is a layer normalization layer.

The output $X_M$ is obtained by applying double-chain constraints to the $M$ layers' features. Finally, the estimated 3D human pose coordinates $\widehat{Y} \in \mathbb{R}^{N \times 3}$ is obtained via the regression head.

\begin{figure}[t]
  \centering
  \includegraphics[width=.9\linewidth]{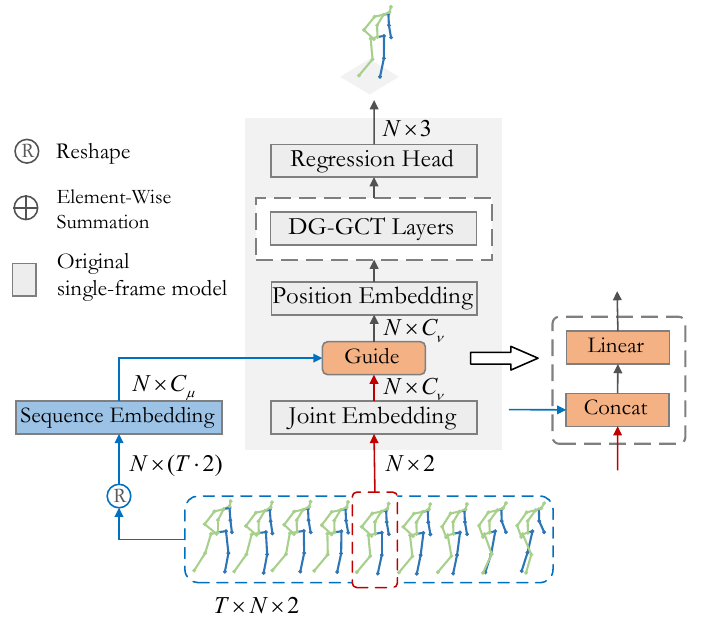}
  \caption{Illustration of the process of DC-GCT in the video domain. The weights trained in a single frame are first loaded, and then the 2D pose sequence and the intermediate frame are used as model inputs to estimate the 3D representation of the intermediate frame.}
  \label{fig:seq}
\end{figure}

\subsection{Extension of Video Sequence Input}

How to effectively input video sequence information into single-frame models is of great research value. Li \textit{et al.} \cite{li2022graphmlp} proposed concatenating the entire time series information of each joint as the embedding for that joint. However, single-frame-based models mainly learn the spatial information of human joints in the current frame, and this method cannot effectively capture the influence of long time series on the current frame. Since the model regresses from 2D video sequences to intermediate frames, as shown in Figure \ref{fig:seq}, we suggest guiding the embedding of the video sequence through the 2D joint embedding of the intermediate frame to capture more important information from long time sequences for the current frame. Specifically, given the input video sequence $\hat{X}\in \mathbb{R}^{T \times N \times 2}$, let $\hat{x}\in \mathbb{R}^{N \times 2}$ be the representation of its intermediate frame. The model guides the embedding of the video sequence through the joint embedding of the intermediate frame, as follows:
\begin{gather}
  \label{eq:guide}
  X_0 = Concat(\hat{X} W_{\mu},\hat{x}W_{\nu})W_{\eta}
\end{gather}
where $W_{\mu}\in \mathbb{R}^{2T \times C_\mu},W_{\nu}\in \mathbb{R}^{2 \times C_\nu},W_{\eta} \in \mathbb{R}^{(C_\mu +C_\nu) \times C_\nu}$ are the linear projection weights. Generally, we set the dimension $C_\mu$ of the sequence embedding to be greater than the dimension $C_\nu$ of the joint embedding. $X_0 \in \mathbb{R}^{N \times C_\nu}$ is the input feature of the single-frame model. In Eq.(\ref{eq:embed}), we add the position embedding to the input feature $X_0$ to obtain the intermediate feature $X_1 \in \mathbb{R}^{N \times C_\nu}$ for the first layer of the double-chain structure. The rest of the model is the same as the single-frame model.

\subsection{Loss Function}

To optimize our DC-GCT, we utilized $L_2$-norm loss to minimize the errors between estimate and ground truth. Additionally, based on the recommendation from \cite{zhang2022mixste}, different weights were assigned to each joint to account for their varying importance. The final loss function can be expressed as follows:
\begin{gather}
\label{eq:lossg}
\mathcal{L} = \frac{1}{N} \sum_{i=1}^N(\theta_i \Vert Y_i - \widehat{Y}_i \Vert_2)
\end{gather}
where $Y_i$ denotes the ground truth 3D joint position for joint $i$, while $\widehat{Y}_i$ represents its estimated counterpart. Whereas $\theta_i$ represents the weighting factor for joint $i$.

\begin{table*}[htbp]
  \huge
\caption{Quantitative comparison with the state-of-the-art methods on Human3.6M under Protocol \#1 and Protocol \#2. The inputs are the detection 2D pose, (SH) denotes the 2D pose detected by Stacked Hourglass network\cite{newell2016stacked}, (CPN) denotes the Cascaded Pyramid Network\cite{chen2018cascaded} and (HRNet) denotes the High-Resolution Net\cite{sun2019deep}. ($\dagger$) adopts the same refinement module as \cite{cai2019exploiting,zou2021modulated}. The best results with CPN are underlined. \textcolor{red}{Red}: best; \textcolor{blue}{Blue}: second best.}
\label{table:1}
\centering
\resizebox{\textwidth}{!}{
\begin{tabular}{lr|ccccccccccccccc|c}
\toprule[1mm]
\bf{Protocol \#1} &                                             & Dir. & Disc. & Eat & Greet & Phone & Photo & Pose & Pur. & Sit & SitD. & Smoke & Wait & WalkD. & Walk & WalkT. & Avg \\
\hline
Martinez \textit{et al.} \cite{martinez2017simple}(SH) &(ICCV'2017) &51.8 &56.2 &58.1 &59.0 &69.5 &78.4 &55.2 &58.1 &74.0 &94.6 &62.3 &59.1 &65.1 &49.5 &52.4 &62.9 \\
Zhao \textit{et al.} \cite{zhao2019semantic}(SH) &(CVPR'2019) &48.2 &60.8 &51.8 &64.0 &64.6 &53.6 &51.1 &67.4 &88.7 &57.7 &73.2 &65.6 &48.9 &64.8 &51.9 &60.8 \\ 
Ci \textit{et al.} \cite{ci2019optimizing}(CPN) &(ICCV'2019) &46.8 &52.3 &44.7 &50.4 &52.9 &68.9 &49.6 &46.4 &60.2 &78.9 &51.2 &50.0 &54.8 &40.4 &43.3 &52.7 \\ 
Liu \textit{et al.} \cite{liu2020comprehensive}(CPN) &(ECCV'2020) &46.3 &52.2 &47.3 &50.7 &55.5 &67.1 &49.2 &46.0 &60.4 &71.1 &51.5 &50.1 &54.5 &40.3 &43.7 &52.4 \\ 
Xu \textit{et al.} \cite{xu2021graph}(CPN) &(CVPR'2021) &45.2 &49.9 &47.5 &50.9 &54.9 &66.1 &48.5 &46.3 &59.7 &71.5 &51.4 &48.6 &53.9 &39.9 &44.1 &51.9 \\ 
Zhao \textit{et al.} \cite{zhao2022graformer}(CPN) &(CVPR'2022) &45.2 &50.8 &48.0 &50.0 &54.9 &65.0 &48.2 &47.1 &60.2 &70.0 &51.6 &48.7 &54.1 &39.7 &43.1 &51.8 \\ 
Cai \textit{et al.} \cite{cai2019exploiting}(CPN)($\dagger$) &(ICCV'2019) &46.5 &48.8 &47.6 &50.9 &52.9 &61.3 &48.3 &45.8 &59.2 &64.4 &51.2 &48.4 &53.5 &39.2 &41.2 &50.6 \\ 

Li \textit{et al.} \cite{li2023pose}(CPN) &(AAAI'2023) &47.9 &50.0 &47.1 &51.3 &51.2 &59.5 &48.7 &46.9 &56.0 &61.9 &51.1 &48.9 &54.3 &40.0 &42.9 &50.5 \\ 

Zeng \textit{et al.} \cite{zeng2020srnet}(CPN) &(ECCV'2020) &44.5 &48.2 &47.1 &47.8 &51.2 &56.8 &50.1 &45.6 &59.9 &66.4 &52.1 &45.3 &54.2 &39.1 &40.3 &49.9 \\ 
Zou \textit{et al.} \cite{zou2021modulated}(CPN)($\dagger$) &(ICCV'2021) &45.4 &49.2 &45.7 &49.4 &50.4 &58.2 &47.9 &46.0 &57.5 &63.0 &49.7 &46.6 &52.2 &38.9 &40.8 &49.4 \\

Zhang \textit{et al.} \cite{zhang2023learning}(CPN)($\dagger$) &(IJCV'2023) &44.8 &49.1 &45.6 &49.3 &51.0 &57.5 &46.7 &45.3 &55.8 &62.5 &51.0 &46.7 &52.8 &38.1 &40.6 &49.1 \\


Cai \textit{et al.} \cite{cai2023htnet}(CPN) &(ICASSP'2023) &44.0 &49.0 &45.2 &47.8 &50.3 &56.6 &47.2 &45.0 &56.3 &65.8 &48.7 &46.0 &52.4 &39.1 &40.2 &48.9 \\

Hassan \textit{et al.} \cite{hassan2023regular}(HRNet)($\dagger$) &(TIP'2023) &41.0 &\textcolor{red}{46.8} &44.0 &48.4 &\textcolor{blue}{47.5} &\textcolor{red}{50.7} &45.4 &\textcolor{blue}{42.3} &\textcolor{blue}{53.6} &65.8 &\textcolor{blue}{45.6} &45.2 &\textcolor{blue}{48.9} &39.7 &40.6 &\textcolor{blue}{47.0}\\

\hline

DC-GCT(CPN) &                        &43.3 &48.1 &44.7 &48.2 &50.3 &58.2 &45.7 &44.8 &56.3 &\textcolor{blue}{62.4} &49.1 &45.4 &52.9 &\textcolor{blue}{37.2} &38.9 &48.4 \\
DC-GCT(CPN)($\dagger$)  &            &\underline{42.2} &\underline{47.3} &\underline{44.6} &\underline{\textcolor{blue}{47.6}} &\underline{49.7} &\underline{56.1} &\underline{\textcolor{blue}{45.3}} &\underline{43.8} &\underline{55.3} &\underline{\textcolor{red}{59.4}} &\underline{48.5} &\underline{\textcolor{red}{44.7}} &\underline{51.0} &\underline{\textcolor{red}{36.8}} &\underline{\textcolor{red}{38.3}} &\underline{47.4} \\

DC-GCT(HRNet) &                        &\textcolor{blue}{42.2} &47.6 &\textcolor{blue}{43.0} &47.7 &47.6 &52.7 &46.1 &42.9 &54.1 &65.2 &46.4 &45.4 &49.2 &38.4 &39.6 &47.2 \\
DC-GCT(HRNet)($\dagger$)  &            &\textcolor{red}{40.6} &\textcolor{blue}{47.0} &\textcolor{red}{42.2} &\textcolor{red}{46.7} &\textcolor{red}{46.8} &\textcolor{blue}{50.8} &\textcolor{red}{44.6} &\textcolor{red}{41.8} &\textcolor{red}{53.2} &63.5 &\textcolor{red}{44.9} &\textcolor{blue}{45.1} &\textcolor{red}{48.0} &38.0 &\textcolor{blue}{38.7} &\textcolor{red}{46.1} \\

\toprule[1mm]


\bf{Protocol \#2} &                              & Dir. & Disc. & Eat & Greet & Phone & Photo & Pose & Pur. & Sit & SitD. & Smoke & Wait & WalkD. & Walk & WalkT. & Avg \\
\hline
Martinez \textit{et al.} \cite{martinez2017simple}(SH) &(ICCV'2017) &39.5 &43.2 &46.4 &47.0 &51.0 &56.0 &41.4 &40.6 &56.5 &69.4 &49.2 &45.0 &49.5 &38.0 &43.1 &47.7 \\
Ci \textit{et al.} \cite{ci2019optimizing}(CPN) &(ICCV'2019) &36.9 &41.6 &38.0 &41.0 &41.9 &51.1 &38.2 &37.6 &49.1 &62.1 &43.1 &39.9 &43.5 &32.2 &37.0 &42.2 \\
Liu \textit{et al.} \cite{liu2020comprehensive} (CPN)&(ECCV'2020) &35.9 &40.0 &38.0 &41.5 &42.5 &51.4 &37.8 &36.0 &48.6 &56.6 &41.8 &38.3 &42.7 &31.7 &36.2 &41.2 \\
Cai \textit{et al.} \cite{cai2019exploiting} (CPN)($\dagger$) &(ICCV'2019) &36.8 &38.7 &38.2 &41.7 &40.7 &46.8 &37.9 &35.6 &47.6 &51.7 &41.3 &36.8 &42.7 &31.0 &34.7 &40.2 \\
Zeng \textit{et al.} \cite{zeng2020srnet} (CPN)&(ECCV'2020) &35.8 &39.2 &36.6 &\underline{ \textcolor{red}{36.9}} &39.8 &45.1 &38.4 &36.9 &47.7 &54.4 &\underline{38.6} &36.3 &\underline{ \textcolor{red}{39.4}} &30.3 &35.4 &39.4 \\
Zou \textit{et al.} \cite{zou2021modulated}(CPN)($\dagger$)  &(ICCV'2021) &35.7 &38.6 &36.3 &40.5 &\underline{39.2} &44.5 &37.0 &35.4 &46.4 &51.2 &40.5 &35.6 &41.7 &30.7 &33.9 &39.1 \\

Cai \textit{et al.} \cite{cai2023htnet}(CPN) &(ICASSP'2023) &35.1 &38.6 &36.6 &39.4 &39.8 &\underline{43.8} &36.7 &34.7 &45.3 &52.8 &39.9 &35.4 &41.8 &31.2 &33.5 &39.0 \\

Hassan \textit{et al.} \cite{hassan2023regular}(HRNet)($\dagger$) &(TIP'2023) &\textcolor{blue}{34.2} &38.2 &35.6 &40.8 &38.5 &\textcolor{blue}{41.8} &36.0 &34.0 &\textcolor{red}{43.9} &56.2 &38.0 &36.3 &40.2 &31.2 &33.3 &38.6 \\


\hline
DC-GCT(CPN)  &                       &\underline{34.4} &\underline{ \textcolor{blue}{37.7}} &35.6 &39.5 &39.5 &44.5 &\underline{ \textcolor{blue}{35.9}} &34.2 &44.9 &\textcolor{blue}{50.1} &39.8 &\underline{ \textcolor{red}{34.7}} &41.5 &\textcolor{blue}{29.5} &\underline{32.3} &38.3 \\ 

DC-GCT(CPN)($\dagger$)  &            &34.7 &37.8 &\underline{35.4} &39.8 &39.4 &44.1 &36.2 &\underline{34.2} &\underline{44.5} &\underline{ \textcolor{red}{48.6}} &39.9 &\textcolor{blue}{34.9} &41.1 &\underline{ \textcolor{red}{29.5}} &32.5 &\underline{ \textcolor{blue}{38.2}} \\

DC-GCT(HRNet) &                        &\textcolor{red}{33.3}  &\textcolor{red}{37.5}  &\textcolor{red}{34.7}  &\textcolor{blue}{39.1}  &\textcolor{red}{38.2}  &\textcolor{red}{41.8}  &\textcolor{red}{35.8}  &\textcolor{red}{33.2}  &\textcolor{blue}{44.0}  &55.1  &\textcolor{blue}{37.9}  &36.1  &\textcolor{blue}{40.0}  &29.9  &\textcolor{red}{31.9}  &\textcolor{red}{37.9}  \\
DC-GCT(HRNet)($\dagger$)  &            &34.7 &38.2 &\textcolor{blue}{34.7} &39.7 &\textcolor{blue}{38.3} &42.1 &36.1 &\textcolor{blue}{33.8} &44.3 &54.6 &\textcolor{red}{37.9} &36.6 &40.6 &30.3 &\textcolor{blue}{32.1} &38.3 \\
\bottomrule[1mm]

\end{tabular}
}
\end{table*}

\begin{figure}[t]
\centering
\includegraphics[width=\linewidth]{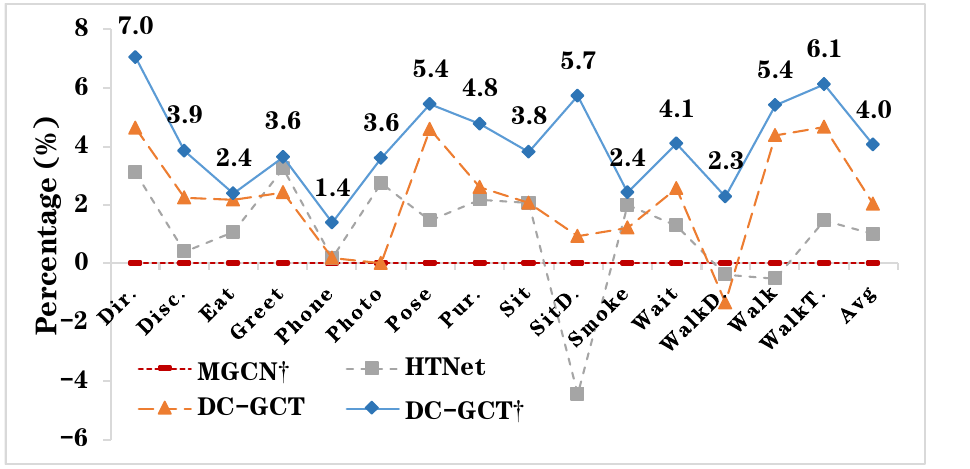}
\caption{Comparison of our proposed method with SOTA methods \cite{zou2021modulated,cai2023htnet} on the test set of the Human3.6M dataset, and evaluated the percentage improvement in human pose estimation outcomes for various actions, utilizing MGCN \cite{zou2021modulated} as a baseline. Using detected 2D poses from Cascaded Pyramid Network (CPN)\cite{chen2018cascaded}. $\dagger$ indicates the use of refinement module.}
\label{fig:chart}
\end{figure}

\begin{table*}[htbp]
  \huge
\caption{Quantitative comparisons of MPJPE in millimeter on Human3.6M under protocol \#1, using ground truth 2D joint locations as input. \textcolor{red}{Red}: best; \textcolor{blue}{Blue}: second best.}
\label{table:2}
\centering
\resizebox{\textwidth}{!}{
\begin{tabular}{lr|ccccccccccccccc|c}
\toprule[1mm]
\bf{Protocol \#1} &                             & Dir. & Disc. & Eat & Greet & Phone & Photo & Pose & Pur. & Sit & SitD. & Smoke & Wait & WalkD. & Walk & WalkT. & Avg \\
\hline
Martinez \textit{et al.} \cite{martinez2017simple} &(ICCV'2017) &37.7 &44.4 &40.3 &42.1 &48.2 &54.9 &44.4 &42.1 &54.6 &58.0 &45.1 &46.4 &47.6 &36.4 &40.4 &45.5 \\
Cai \textit{et al.} \cite{cai2019exploiting} &(ICCV'2019) &33.4 &39.0 &33.8 &37.0 &38.1 &47.3 &39.5 &37.3 &43.2 &46.2 &37.7 &38.0 &38.6 &30.4 &32.1 &38.1 \\ 
Liu \textit{et al.} \cite{liu2020comprehensive} &(ECCV'2020) &36.8 &40.3 &33.0 &36.3 &37.5 &45.0 &39.7 &34.9 &40.3 &47.7 &37.4 &38.5 &38.6 &29.6 &32.0 &37.8 \\ 
Zeng \textit{et al.} \cite{zeng2020srnet} &(ECCV'2020) &35.9 &36.7 &29.3 &34.5 &36.0 &42.8 &37.7 &31.7 &40.1 &44.3 &35.8 &37.2 &36.2 &33.7 &34.0 &36.4 \\ 
Xu \textit{et al.} \cite{xu2021graph} &(CVPR'2021) &35.8 &38.1 &31.0 &35.3 &35.8 &43.2 &37.3 &31.7 &38.4 &45.5 &35.4 &36.7 &36.8 &27.9 &30.7 &35.8 \\ 
Zhao \textit{et al.} \cite{zhao2022graformer} &(CVPR'2022) &\textcolor{blue}{32.0} &38.0 &30.4 &34.4 &\textcolor{blue}{34.7} &43.3 &\textcolor{red}{35.2} &31.4 &38.0 &46.2 &34.2 &35.7 &36.1 &27.4 &30.6 &35.2 \\

Zhang \textit{et al.} \cite{zhang2023learning} &(IJCV'2023) &32.4 &\textcolor{blue}{36.5} &30.1 &\textcolor{blue}{33.3} &36.3 &43.5 &36.1 &\textcolor{blue}{30.5} &37.5 &45.3 &\textcolor{blue}{33.8} &35.1 &35.3 &27.5 &30.2 &34.9 \\

Li \textit{et al.} \cite{li2023pose} &(AAAI'2023) &32.9 &38.3 &\textcolor{blue}{28.3} &33.8 &34.9 &\textcolor{blue}{38.7} &37.2 &30.7 &\textcolor{red}{34.5} &\textcolor{blue}{39.7} &33.9 &\textcolor{blue}{34.7} &\textcolor{blue}{34.3} &\textcolor{red}{26.1} &\textcolor{blue}{28.9} &\textcolor{blue}{33.8} \\ 

\hline
DC-GCT   &        &\textcolor{red}{30.7} &\textcolor{red}{35.6} &\textcolor{red}{26.8} &\textcolor{red}{32.0} &\textcolor{red}{32.6} &\textcolor{red}{36.7} &\textcolor{blue}{36.0} &\textcolor{red}{30.0} &\textcolor{blue}{35.1} &\textcolor{red}{36.5} &\textcolor{red}{32.2} &\textcolor{red}{33.6} &\textcolor{red}{33.9} &\textcolor{blue}{26.5} &\textcolor{red}{27.5} &\textcolor{red}{32.4} \\
\bottomrule[1mm]

\end{tabular}
}
\end{table*}

\begin{figure*}[t]
\centering
\includegraphics[width=.8\textwidth]{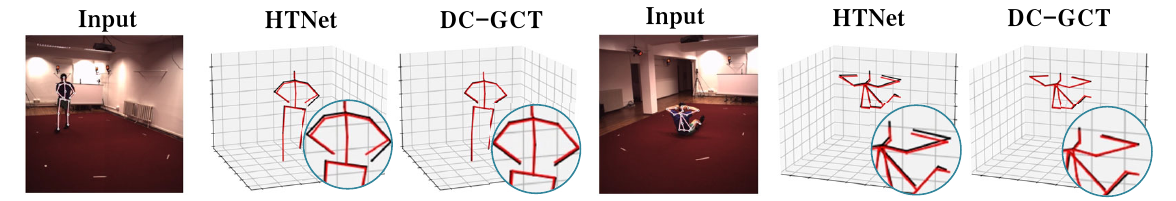}
\caption{Qualitative comparison among the proposed method (DC-GCT) and the previous state-of-the-art method (HTNet) \cite{cai2023htnet}  on Human3.6M dataset. The circles indicate locations where our model estimates result in better results.}
\label{fig:show}
\end{figure*}

\section{Experiments}

\subsection{Datasets and Evaluation Protocols}

We evaluate the proposed model on two 3D human pose estimation datasets: Human3.6M \cite{ionescu2013human3} and MPI-INF-3DHP \cite{mehta2017monocular}.

{\bf{Human3.6M}} is a widely used dataset comprising 3.6 million video frames and annotations for 3D human pose estimation. The dataset features 15 activities performed by 11 actors, captured by four cameras from different angles. Following the same strategy of the previous methods \cite{zou2021modulated,cai2023htnet}, the model is trained on five subjects (S1, S5, S6, S7, S8) and evaluated on two subjects (S9, S11) on a 17-joint skeleton.

{\bf{MPI-INF-3DHP}} is a challenging benchmark dataset that contains 1.3 million frames of 3D pose data, exhibiting more diverse motions than Human3.6M. It comprises three different scenarios: green screen (GS), non-green screen(noGS), and outdoor (Outdoor). To quantitatively demonstrate the model's generalization ability, we tested our model trained on Human3.6M on the test set of MPI-INF-3DHP.

{\bf{Evaluation Protocols.}} For Human3.6M, We use the most commonly used evaluation metrics (MPJPE and PMPJPE). MPJPE (\textit{i.e.} Protocol \#1) is to measure the mean Euclidean distance between predicted and ground-truth joint positions without any transformation. P-MPJPE (\textit{i.e.} Protocol \#2) is the MPJPE after aligning the predicted 3D pose with the ground-truth using a rigid transformation. For MPI-INF-3DHP, we follow previous works \cite{zhao2019semantic,zou2021modulated} and report metrics of Percentage of Correct Keypoint(PCK) within 150mm range, and Area Under Curve(AUC) for a range of PCK thresholds  as evaluation metrics in our experiments.

\subsection{Implementation Details}
Following previous work\cite{cai2019exploiting,zou2021modulated,cai2023htnet,hassan2023regular}, we obtain 2D pose detections using the cascaded pyramid network (CPN)\cite{chen2018cascaded} and the high-resolution net (HRNet)\cite{sun2019deep}. We implemented our method using Pytorch \cite{paszke2019pytorch} and trained it on a single NVIDIA GeForce RTX 2080 SUPER. The model was trained for 30 epochs with a batch size of 512. The initial learning rate was 0.0005 for single-frame input and 0.001 for video sequence input, and a decay factor of 0.95 was applied after each epoch, with a decay rate of 0.5 every five epochs. DC-GCT consists of three stacked double-chain structures, where the input channel dimension is 160 and the ratio of double-chain's dimensions is 1:4. We employed refinement modules similar to \cite{cai2019exploiting,zou2021modulated}. For data augmentation, we only applied horizontal flip data augmentation.

\subsection{Comparison with State-of-the-art}

{\bf{Comparison on Human3.6M.}} We compare our DC-GCT with previous state-of-the-art methods on the Human3.6m dataset. As shown in Table \ref{table:1}, we conduct experiments using 2D poses obtained as input from two detectors, a typical 2D detector CPN\cite{chen2018cascaded} used in previous methods, and a more advanced 2D detector HRNet\cite{sun2019deep} that provides more accurate 2D poses. For CPN detector, our method achieves 48.4 mm under MPJPE and 38.3mm under PA-MPJPE, outperforming all previous methods. Note that some works \cite{cai2019exploiting,zou2021modulated} adopt a pose refinement module for further performance improvement. Compared with them, DC-GCT achieves lower MPJPE without refinement module. Furthermore, we achieve 47.4 mm under MPJPE by incorporating such a module, which reduces the error by 2 mm (relative 4\% improvement) compared to MGCN \cite{zou2021modulated}. Notably, our model outperforms the previous SOTA methods for all action categories, which is also shown in Figure \ref{fig:chart}.

For HRNet detector, due to the higher accuracy of the 2D poses, the performance of our methods is improved. Our method achieves 47.2 mm under MPJPE and 37.9mm under PA-MPJPE, which is 1.2mm and 0.4mm lower than the previous SOTA method of CPN detector, respectively. With the refinement module, we achieve 46.1 mm under MPJPE, outperforming all previous SOTA methods.

Moreover, as shown in Table \ref{table:2}, with ground truth 2D keypoints as input, DC-GCT achieves the best performance with 32.4 mm MPJPE, representing a 2.8mm decrease in error (relative 8\% improvement) over the SOTA method. We also achieve first place in 12 of the 15 action categories.

We provide some qualitative results in Figure \ref{fig:show}. We compare our proposed method with HTNet \cite{cai2023htnet} on the Human3.6M dataset. Our method predicts more accurate 3D poses.

\begin{table}[t]
\caption{Quantitative comparison on MPI-INF-3DHP. GS denotes green screen. \textcolor{red}{Red}: best; \textcolor{blue}{Blue}: second best.}
\label{table:3}
\centering
\resizebox{\linewidth}{!}{
\begin{tabular}{l|c|c|c|c|c}
\toprule[.3mm]
\multirow{2}{*}{Method} &\multicolumn{4}{c|}{PCK} &\multirow{2}{*}{AUC$\uparrow$} \\
\cline{2-5}

& GS$\uparrow$ & noGS$\uparrow$ & Outdoor$\uparrow$ & ALL$\uparrow$ & \\
\hline
Martinez \textit{et al.} \cite{martinez2017simple}  &49.8 &42.5 &31.2 &42.5 &17.0 \\
Ci \textit{et al.} \cite{ci2019optimizing}          &74.8 &70.8 &77.3 &74.0 &36.7 \\ 
Zeng \textit{et al.} \cite{zeng2020srnet}           &- &- &80.3 &77.6 &43.8 \\ 
Li \textit{et al.} \cite{li2019generating}          &70.1 &68.2 &66.6 &66.9 &- \\ 
Zhao \textit{et al.} \cite{zhao2022graformer}       &80.1 &77.9 &74.1 &79.0 &43.8 \\
Liu \textit{et al.} \cite{liu2020comprehensive}     &77.6 &80.5 &80.1 &79.3 &47.6 \\ 
Xu \textit{et al.} \cite{xu2021graph}               &81.5 &81.7 &75.2 &80.1 &45.8 \\ 
Zou \textit{et al.} \cite{zou2021modulated}         &86.4 &86.0 &85.7 &86.1 &53.7 \\

Cai \textit{et al.} \cite{cai2023htnet}         &\textcolor{blue}{86.9} &\textcolor{blue}{86.2} &\textcolor{blue}{85.9} &\textcolor{blue}{86.7} &\textcolor{blue}{54.1} \\
\hline
DC-GCT(Ours)                                        &\textcolor{red}{88.5} &\textcolor{red}{87.4} &\textcolor{red}{86.1} &\textcolor{red}{87.5} &\textcolor{red}{55.9} \\

\bottomrule[.3mm]

\end{tabular}
}
\end{table}

{\bf{Comparison on MPI-INF-3DHP.}} As shown in Table \ref{table:3}, our proposed DC-GCT method was compared with previous state-of-the-art methods in cross-dataset evaluation scenarios, where we trained only on the Human3.6M training dataset and tested on the MPI-INF-3dHP testing dataset. The results showed that our method consistently outperformed all other methods across all evaluation scenarios and performance metrics, demonstrating its strong ability to generalize to unknown environments.

\begin{table}[t]
\caption{Quantitative comparison with temporal methods on Human3.6M. \textcolor{red}{Red}: best; \textcolor{blue}{Blue}: second best.}
\label{table:temporal}
\centering
\resizebox{.8\linewidth}{!}{
\begin{tabular}{l|c|c|c}
\toprule
Method &Frame &Params &MPJPE($mm$)\\
\hline
Pavllo \textit{et al.} \cite{pavllo20193d}  &9 &\textcolor{blue}{4.4M} &49.8 \\
Cai  \textit{et al.} \cite{cai2019exploiting}  &7 &5.1M &48.8 \\
Zheng  \textit{et al.} \cite{zheng20213d}  &9 &9.6M &49.9 \\
Li  \textit{et al.} \cite{li2022mhformer}  &9 &18.9M &\textcolor{blue}{47.8} \\

\hline
DC-GCT(Ours) &9 &\textcolor{red}{2.28M} &\textcolor{red}{46.46} \\
\hline

Pavllo  \textit{et al.} \cite{pavllo20193d}  &27 &8.6M &48.8 \\
Liu  \textit{et al.} \cite{liu2020attention}  &27 &\textcolor{blue}{5.7M} &48.5 \\
Zheng  \textit{et al.} \cite{zheng20213d}  &27 &9.6M &47.0 \\
Li  \textit{et al.} \cite{li2022mhformer}  &27 &18.9M &\textcolor{blue}{45.9} \\
\hline
DC-GCT(Ours) &27 &\textcolor{red}{2.32M} &\textcolor{red}{45.69} \\
\bottomrule

\end{tabular}
}
\end{table}

\begin{figure}[t]
  \centering
  \includegraphics[width=.9\linewidth]{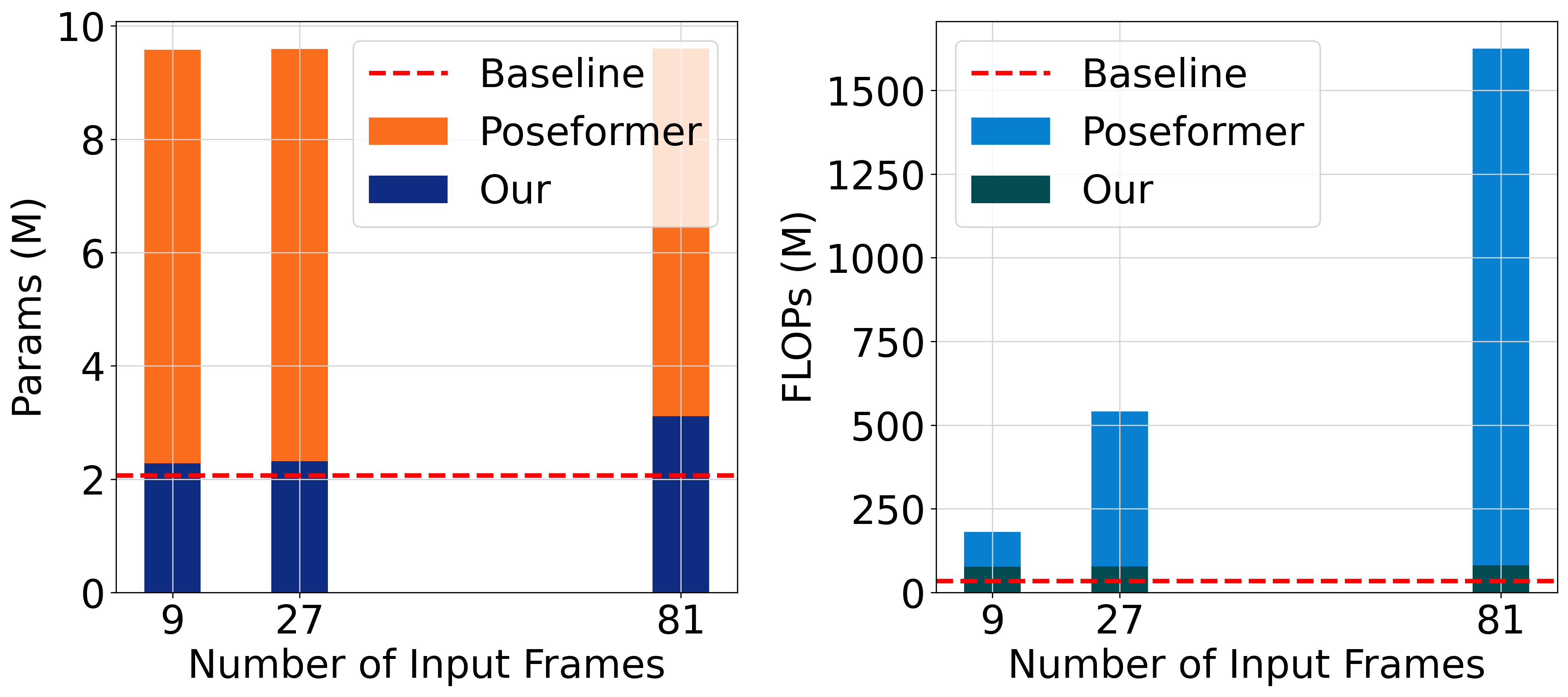}
  \caption{Comparison of Params and FLOPs of our method with Poseformer\cite{zheng20213d} under different input frames. Where baseline is the performance of our model on a single frame of input.}
  \label{fig:params}
\end{figure}

{\bf{Comparison with Temporal Methods.}} To explore the portability of DC-GCT in video-based 3D human pose estimation, we compared several methods \cite{pavllo20193d,cai2019exploiting,zheng20213d,li2022mhformer,liu2020attention} using similar frames as inputs. As shown in Table \ref{table:temporal}, DC-GCT (C=160, M=3) achieves 46.46mm with 9 input frames and 45.69mm with 27 input frames under MPJPE, with only 2.28M and 2.32M model parameters, which is significantly fewer than other methods. The experiments demonstrate that our approach can achieve competitive performance with fewer parameters in multi-frame-based 3D human pose estimation. 

Additionally, we compared the params and FLOPs of our method with Poseformer \cite{zheng20213d} under different input frames, as shown in Figure \ref{fig:params}. With the increase in the number of input frames, the parameter count of Posformer experiences a relatively minor increment, yet there is a significant surge in FLOPs which, in contrast, exhibits a negligible rise in the case of our approach. The latter's marginal increase in FLOPs can be disregarded, rendering it comparable to the computational load of the baseline model operating with single-frame inputs. This observation underscores the capacity of our single-frame model to maintain a low level of computational complexity when faced with an augmentation of input frames, in stark contrast to the Poseformer model. Consequently, this circumstance endows our single-frame model with a certain advantage in processing multi-frame inputs.


\subsection{Ablation Study}

To evaluate the validity of the proposed model, we conducted ablation experiments on the Human3.6m dataset using Protocol \#1, with CPN \cite{chen2018cascaded} as the 2D detector.

\begin{table}[ht]
  \Huge
\caption{Ablation study on different components in DC-GCT. L2G denotes the local-to-global chain, G2L denotes the global-to-local chain, (1:4) denotes the channel dimension ratio of the two chains.}
\label{table:architecture}
\centering
\resizebox{.9\linewidth}{!}{
\begin{tabular}{c|c|c|c|c|c|c|c|c}
\toprule[1mm]

\multirow{2}{*}{} &\multicolumn{2}{c|}{L2G} & \multicolumn{2}{c|}{G2L} & &\multirow{2}{*}{\makecell[c]{Ratio \\ (1:4)}} &\multirow{2}{*}{Params} &\multirow{2}{*}{MPJPE($mm$)}\\
\cline{2-5}
&LCM &GCM &GCM &LCM & FIM & & & \\
\hline
\multirow{4}{*}{Single-chain}&\Checkmark &  &  &  & & &2.25M &52.51\\
& & \Checkmark &  &  & & &0.93M &52.89\\
&\Checkmark & \Checkmark &  &  &  &  &2.56M &49.45\\
&  &  & \Checkmark & \Checkmark &   &  &2.56M &49.49\\
\hline
\multirow{4}{*}{\makecell[c]{Double-chain}}& \Checkmark &  & \Checkmark &   &  &  &1.11M &49.81\\
& \Checkmark &  \Checkmark & \Checkmark & \Checkmark &   &  &1.56M &49.38\\
& \Checkmark &  \Checkmark & \Checkmark & \Checkmark & \Checkmark &  &1.68M &48.79\\
& \Checkmark &  \Checkmark & \Checkmark & \Checkmark & \Checkmark & \Checkmark &2.07M &\bf{48.41}\\
\bottomrule[1mm]

\end{tabular}
}
\end{table}

{\bf{Model Components.}} We investigated the impact of various component combinations on model results, as presented in Table \ref{table:architecture}. For the single-chain design, using solely LCM or GCM resulted in MPJPE of 52.51mm and 52.89mm, respectively. However, both LCM based on GCN and GCM based on self-attention mechanisms have limitations in effectively capturing the human body's dependency relationships. In contrast, the single-chain dual-constraint designs have significantly improved performance despite increased parameters, such as local-to-global (L2G) single-chain and global-to-local (G2L) single-chain designs with MPJPE of 49.45mm and 49.49mm, respectively. However, the transition phase in these designs has the potential to cause information loss, such as when applying local-to-global constraints where the information flow may lose global dependency information after passing through the local constraints. Applying global-to-local constraints may cause structural and local dependency information loss. For the double-chain design, we applied local and global constraints in parallel by LCM and GCM, respectively. However, the significant difference between the two chains made effective integration of information challenging, resulting in an MPJPE of 49.81mm. To address this issue, we designed a local-to-global and global-to-local double-chain structure to reduce the difference between the two chains, resulting in a slight improvement in MPJPE to 49.38mm, but information loss during the transition phase remained an issue. Subsequently, we introduced FIM to the transitional phase of the constraint to automatically fuse and map local and global features, reducing information loss and achieving an MPJPE of 48.79mm. Finally, we applied the channel dimension ratios of the two chains to 1:4, resulting in an MPJPE of 48.41mm.

\begin{table}[ht]
\caption{Ablation study on the hyperparameters of our model. $M$ denotes the number of layers, $C$ denotes the channel dimension.}
\label{table:parameter}
\centering
\resizebox{.8\linewidth}{!}{
\begin{tabular}{c|c|c|c|c}

\toprule
&\multicolumn{4}{c}{$C=160$}\\
\cline{2-5}
$M$&1 &2 &3 &4 \\
\hline
Params &0.66M &1.38M &2.07M &2.76M \\
\hline
MPJPE($mm$) &52.31 &49.39 &\bf{48.41} &48.57 \\
\midrule
&\multicolumn{4}{c}{$M=3$}\\
\cline{2-5}
$C$&80 &160 &200 &320 \\
\hline
Params &0.52M &2.07M &3.23M &8.23M \\
\hline
MPJPE($mm$) &50.04 &\bf{48.41} &48.93 &49.35 \\
\bottomrule
\end{tabular}
}
\end{table}

{\bf{Model Hyperparameters.}} This section investigates several hyperparameter combinations to determine the optimal network architecture in Table \ref{table:parameter}. The parameters examined are the number of layers ($M$), channel dimension ($C$), and their effect on the number of parameters and MPJPE.In the case where the channel dimension of the model is 160, increasing the number of layers leads to an increase in the number of parameters. As $M$ increases from 1 to 4, the number of parameters increases from 0.66M to 2.76M, while MPJPE reaches performance saturation at $M=3$, with a value of 48.41mm. Since our GCM adopts an 8-headed multi-head attention mechanism and our double-chain dimension ratio is 1:4, we set the total channel dimension to be a multiple of 40 in our experiments. Therefore, when the number of layers is fixed at 3, we compared the performance of $C$ at 80, 160, 200, and 320. It can be observed that when $C=160$, the number of parameters is 2.07M and MPJPE is 48.41mm, which is optimal. Consequently, the optimal hyperparameter combination for our model in this task is when both the number of layers and channel dimension are set to 3 and 160.

\begin{table}[ht]
  \renewcommand{\arraystretch}{1.2}
\caption{Ablation study on different double-chain configurations. $C_1$ and $C_2$ denote the channel dimension of local-to-global chain and global-to-local chain of the double-chain, respectively. Ratio denotes the ratio of $C_1$ and $C_2$.}
\label{table:channel}
\centering
\begin{tabular}{l|c|cc|c|c}
\toprule
Ratio  &$C$ &$C_1$ &$C_2$   &Params &MPJPE($mm$)\\
\hline
1:1   &160  &80   &80        &1.68M   &48.79 \\
1:3   &160  &40   &120       &1.96M   &48.72\\
3:1   &160  &120  &40     &1.96M   &48.96 \\
4:1   &160  &128  &32     &2.07M   &48.92 \\
1:5   &192 &32   &160    &3.10M &48.91 \\
5:1   &192 &160   &32    &3.10M &49.13 \\
\hline
1:4   &160 &32   &128    &2.07M &\bf{48.41} \\
\bottomrule

\end{tabular}
\end{table}

{\bf{Double-chain Configurations.}} We also investigated the influence of different double-chain configurations on performance. As our model is a double-chain structure, the total channel dimension of the model is divided into two parts: the channel dimension of local-to-global chain and the channel dimension of global-to-local chain. By adjusting the ratio of $C_1$ and $C_2$, the total channel dimension and parameter quantity of the model can be controlled. From Table \ref{table:channel}, it can be seen that when the total channel dimension is 160, the ratio of $C_1$ to $C_2$ can be 1:1, 1:3, 3:1, 1:4, 4:1, and the greater the difference between $C_1$ and $C_2$, the larger the parameter quantity of the model. When the ratio is 1:4, the MPJPE reaches the minimum value of 48.41mm with 2.07M parameters. This section also explores the case where $C_1$ and $C_2$ have a greater difference, namely 1:5 and 5:1. At this time, the total channel dimension of the model increases to 192, and the corresponding parameters also increase, but the MPJPE are not as good as that of the 1:4 and 4:1 case, which are 48.91mm and 49.13mm.


\begin{table}[htbp]
  \renewcommand{\arraystretch}{1.2}
\caption{Ablation study on the different input frames on the Human3.6m dataset.}
\label{table:temporal_1}
\centering
\resizebox{.9\linewidth}{!}{
\begin{tabular}{c|c|c|c|c}
\toprule
&9 &27 &81 &243\\
\hline
Params &2.28M &2.32M &3.11M &3.44M\\
\hline
FLOPs(M)&77 &78 &82 &93\\
\hline
MPJPE(mm)&46.46 &45.69 &\bf{44.73} &44.87\\
\bottomrule
\end{tabular}
}
\end{table}

{\bf{Input Frames.}} We also investigated the influence of different input frames on performance. As shown in Table \ref{table:temporal_1}, we compared the performance of our model with different input frames. When the number of input frames was 81 and 243, we used different configurations where M=4, C=160, while the setting with fewer frames maintained the M=3, C=160 configuration. As the number of input frames increased, the model parameter size and computational cost showed an increasing trend. When the number of input frames was 81, the model parameter size was 3.11M and the computational cost was 82M FLOPs, achieving the lowest pose estimation error MPJPE of 44.73mm. Further increasing the input frames to 243 frames increased the model parameter size and computational cost, but did not continue to decrease the pose estimation error, implying that the model performance was saturated. 


\begin{figure}[ht]
  \centering
  \includegraphics[width=\linewidth]{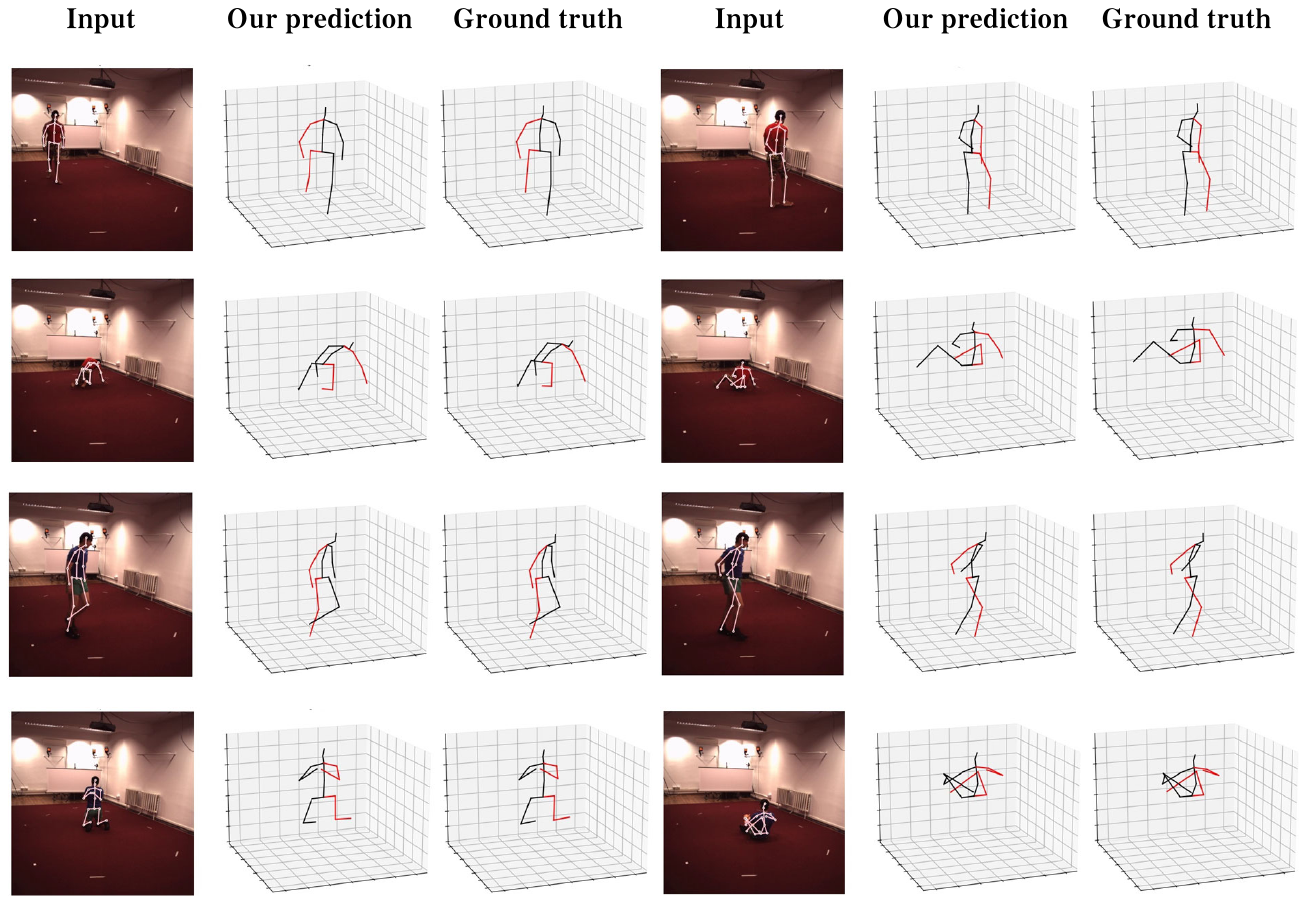}
  \caption{Qualitative visual results of our method on the Human3.6M test set. }
  \label{fig:show1}
\end{figure}

\begin{figure}[ht]
  \centering
  \includegraphics[width=\linewidth]{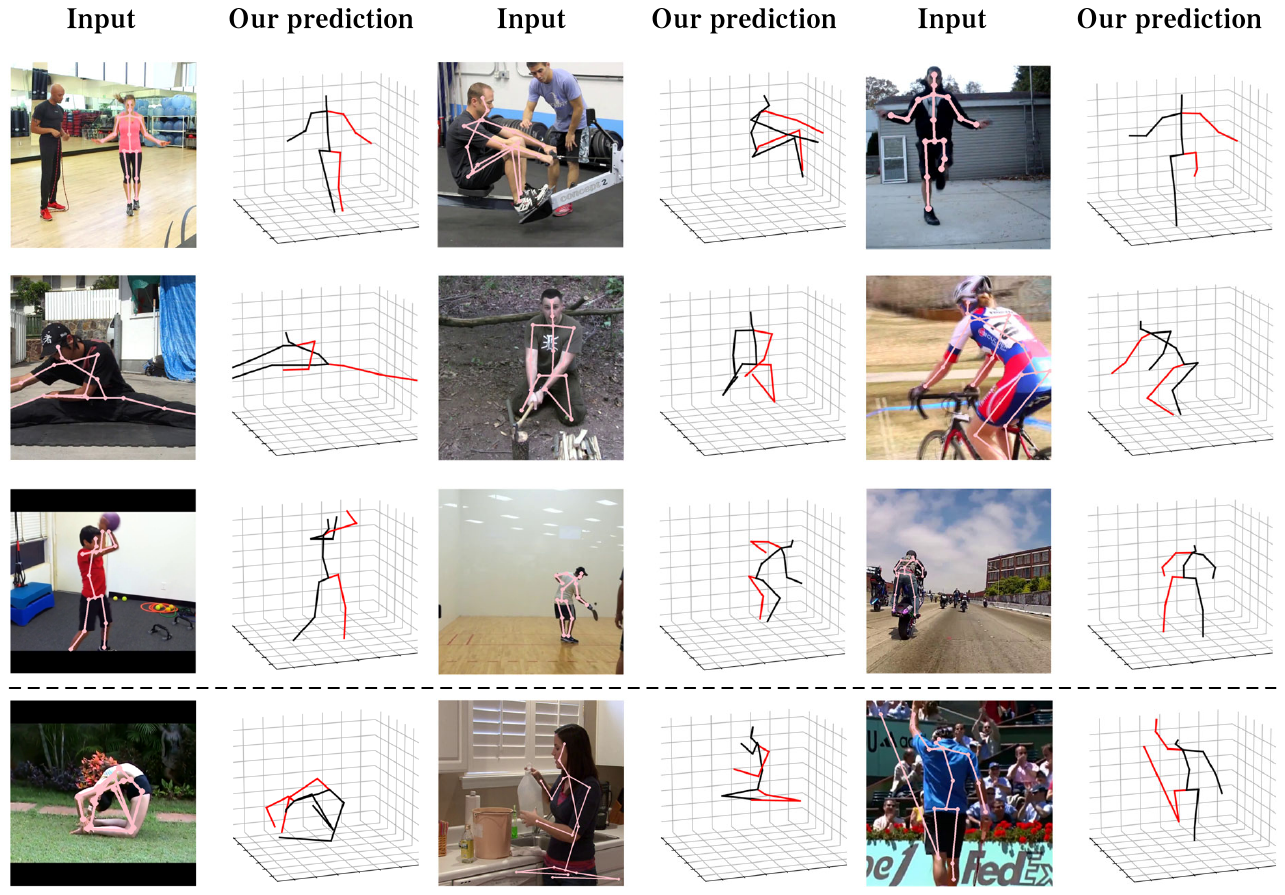}
  \caption{Visualization results of the proposed DC-GCT on challenging in-the-wild images. The bottom row shows three 3D human poses that cannot be accurately generated.}
  \label{fig:show3}
\end{figure}

\subsection{Qualitative Results}

Figure \ref{fig:show1} shows a visual comparison between the predicted and ground truth poses. The DC-GCT was evaluated with Walking and SittingDown actions on the Human3.6M test sets S9 and S11, and our model accurately predicted the 3D human pose for indoor environments. To evaluate its generalization performance, Figure \ref{fig:show3} shows the visualization results of DC-GCT on MPII \cite{andriluka20142d} dataset that contains in-the-wild images. Our proposed model can accurately predict actions that differ from the Human3.6M dataset in most in-the-wild scenes. However, as shown in the bottom row of Figure \ref{fig:show3}, there are still some failure cases due to heavy occlusion, half-body, 2D detection errors, and other factors that cannot be accurately predicted by our method.

\section{Conclusion}

This paper proposes a novel model, DC-GCT, for 3D human pose estimation from monocular images or videos. We combine the advantages of GCN and Transformers and design LCM and GCM modules to impose local and global constraints, respectively. And we design a FIM module to enable feature interaction. By combining these three modules, our model achieves double-chain constraints to capture the multi-level dependencies between human joints. Comprehensive evaluations show that our DC-GCT outperforms state-of-the-art techniques in 3D human pose estimation tasks. Especially in the evaluation of Human3.6M dataset using detected 2D poses from CPN, our results achieve optimality in all action categories. Our proposal presents a flexible and scalable double-chain architecture that can incorporate advanced constraints and interaction modules in future research. Moreover, we provide an efficient method for embedding video sequences into single-frame models. This research provides valuable insights to further advance the development of related technologies in this field, contributing to the broader goal of enhancing human-centric applications such as virtual reality, gaming, and human computer interaction.

\normalem
\bibliographystyle{IEEEtran}
\bibliography{IEEEabrv,reference}


\vfill

\end{document}